Perspective

# Building artificial intelligence virtual tissue (AIVT) for tissue state representation, feature prediction, and dynamic simulation

Qiqi Lu[1,2,3], Qianjin Feng[1,2,3], Shaoqun Zeng[4] & Shenghua Cheng[1,2,3,*]

[1]School of Biomedical Engineering, Southern Medical University, Guangzhou, China.
[2]Guangdong Provincial Key Laboratory of Medical Image Processing, Southern Medical University, Guangzhou, China.
[3]Guangdong Province Engineering Laboratory for Medical Imaging and Diagnostic Technology, Southern Medical University, Guangzhou, China.
[4]MOE Key Laboratory for Biomedical Photonics, Wuhan National Laboratory for Optoelectronics, Huazhong University of Science and Technology, Wuhan, China.

*Correspondence: chengsh2023@smu.edu.cn

## Abstract

Modeling tissue states and their transitions is essential for understanding tissue homeostasis in health and pathological remodeling in disease. However, conventional computational modeling approaches are inadequate to capture the complexity of tissues as spatially organized, multiscale biological systems. Artificial intelligence (AI) has shown a remarkable ability for representing intricate systems, creating new opportunities to characterize tissue states and their transitions. Here, we propose the concept of AI virtual tissue (AIVT), an AI framework grounded in spatial multimodal data for modeling tissues in health and disease. AIVT is designed to learn unified, spatially resolved, and dynamically manipulatable representations of tissue state, enabling tissue state representation and analysis, molecular and morphological feature prediction, and simulation of spatiotemporal tissue dynamics. We outline the fundamental assumptions, core capabilities, architectural components, as well as data and algorithm foundations of AIVT as a framework for AI-driven tissue modeling.

## Introduction

Biological function is intimately linked to tissue state, which encompasses molecular composition, cellular organization, tissue architecture, and functional properties. During development, tissues undergo continuous remodeling, accompanied by coordinated changes in their structure and composition[1]. Infectious and inflammatory processes disrupt tissue homeostasis, leading to edema, immune cell infiltration, and localized tissue damage[2]. In cancer, tissue heterogeneity and the spatial organization of tumor microenvironment shape tumor progression[3] and therapeutic response. Therefore, elucidating biological processes in health and disease requires investigation of tissue-level spatial characteristics and spatiotemporal dynamics[4].

To investigate tissue function and behavior, conventional computational tissue models translate biological observations into mathematical rules for simulating the states of individual cells and their dynamic interactions within tissue environments[5–7]. These rules are explicitly defined on the basis of observations, prior knowledge, or hypothesis[5], enabling researchers to examine biological mechanisms ranging from intracellular signaling to collective cell behavior[8]. However, such models are typically tailored to relatively specific questions[9], and their reliance on incomplete prior knowledge and simplified mathematical formulations limits their ability to fully capture the complexity of tissues as spatially organized, multiscale biological systems.

Recent breakthroughs in spatial multimodal technologies have enabled tissue profiling across multiple scales (Fig. 1), from molecular measurements[10], such as genomics, transcriptomics, proteomics, and metabolomics, to morphological imaging[11], including conventional histology, open-top light-sheet (OTLS) microscopy[12,13], micro-computed tomography (microCT)[14], and electron microscopy (EM)[15]. Spatial transcriptomic techniques, such as HDST[16], Visium HD[17], and Stereo-seq[18,19], enable whole-transcriptome mapping at single-cell resolution, while spatial proteomic techniques can simultaneously profile dozens[20] to thousands[21,22] of proteins within a single tissue section. Beyond the increasing depth and breadth of individual modalities, spatially aligned integration of complementary technologies provides a more comprehensive view of tissue complexity[23], including the co-profiling of transcriptome and morphology[24,25], transcriptome and proteome[26,27], transcriptome and epigenome[28–30], proteome and metabolome[31], and transcriptome and metabolome[32]. Such multiscale spatial profiling offers unprecedented opportunities to characterize tissue states with spatial resolution, while also posing substantial challenges for computational tissue modeling.

Artificial intelligence (AI) has demonstrated powerful capabilities for modeling complex systems. For example, when combined with omics, AI techniques have fueled efforts to represent and simulate the behavior of cells[33,34] and organoids[35]. Together with advances in spatial multimodal technologies[36], AI now offers new opportunities to characterize tissue states and their transitions. Here, we propose the concept of AI virtual tissue (AIVT), an AI framework for modeling tissues in health and disease. Distinct from conventional multimodal integration approaches that align, fuse or translate heterogeneous measurements[37–39], AIVT aims to learn unified, spatially resolved, and dynamically manipulatable representations of tissue states from spatial multimodal data for comprehensive tissue-level modeling. AIVT is expected to enable tissue state representation and analysis, molecular and morphological feature prediction, and simulation of spatiotemporal tissue dynamics (Fig. 1), with the potential to advance our understanding of tissue biology and support studies of complex diseases, precision medicine, and therapeutic strategies.

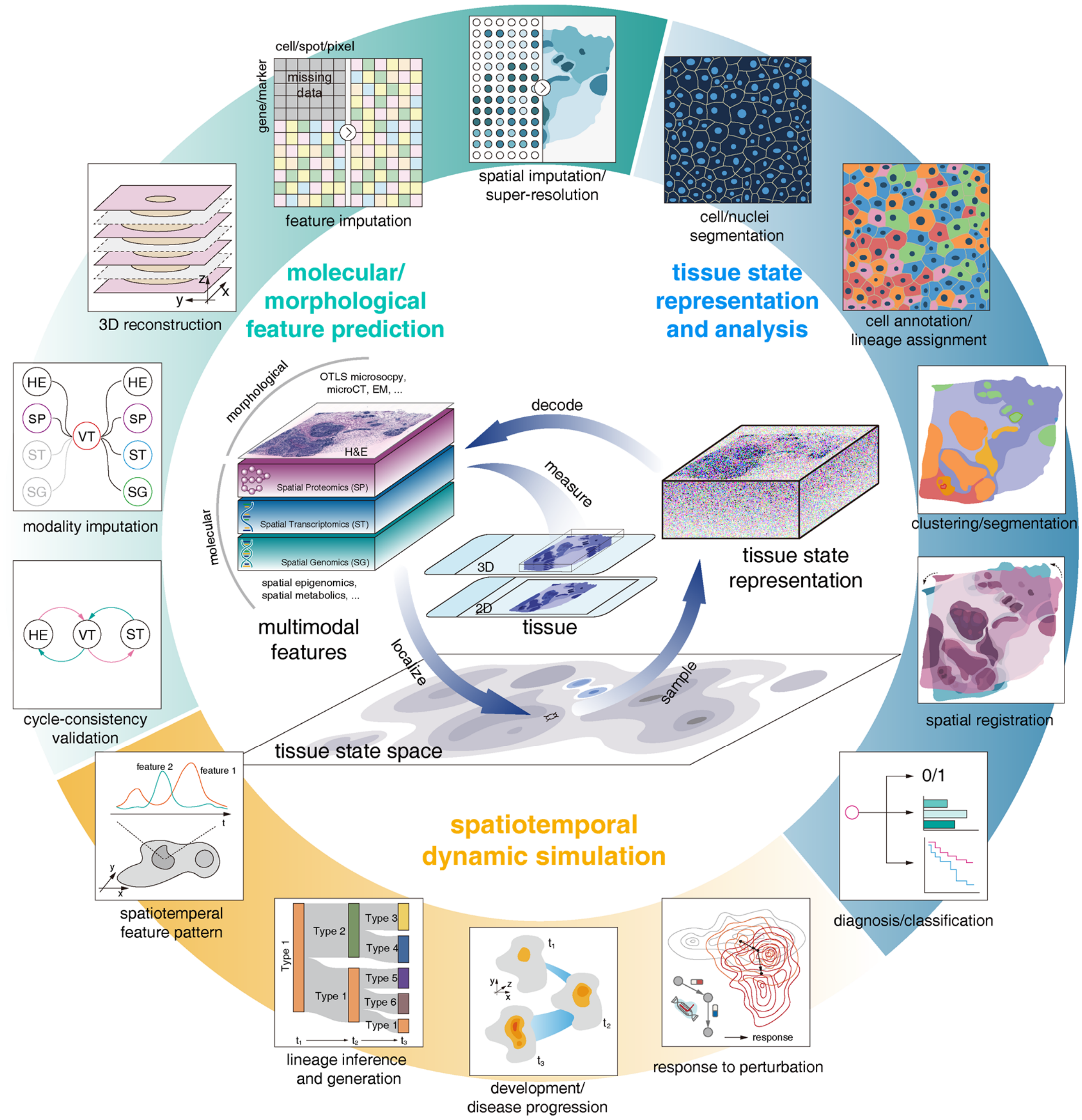

**Fig. 1 | Concept and capabilities of AIVT.** AIVT leverages multimodal measurements to localize tissue states within a tissue state space, and the resulting tissue state representations can be decoded into human-interpretable multimodal features. AIVT can support (1) multi-scale analysis of tissues based on tissue state representations, ranging from cell segmentation and annotation, to spatial clustering, tissue segmentation and registration, and tissue-level disease classification, diagnosis and prognosis assessment; (2) molecular and morphological feature prediction, such as imputation of missing modalities, unmeasured genes, protein markers, and other biological features, three-dimensional volume reconstruction, and cycle-consistency-based validation of results; and (3) simulation of spatiotemporal tissue dynamics, including tissue state responses to perturbations, tissue evolution during development and disease progression, lineage inference and generation, and spatiotemporal feature pattern discovery.

## Fundamental assumptions and core capabilities of AIVT

The concept of AIVT is built on four fundamental assumptions that guide its design and functionality. First, we assume that tissue states can be represented in a latent tissue state space (Fig. 1). Each position in this high-dimensional space corresponds to a specific state of a tissue entity (e.g., a whole tissue sample), and encodes its molecular composition, morphological characteristics, spatial organization, and functional properties. Tissue state transitions associated with homeostasis maintenance[40], development, disease progression[4], and responses to perturbations[41] can therefore be represented as movements within this tissue state space. Second, for the same tissue entity, different modalities capture distinct yet complementary aspects of its state and therefore provide modality-specific information. At the same time, the shared underlying biological identity implies that they contain common information. This complementarity has motivated the development of multimodal integration methods[38,42,43], while the shared biological identity provides a foundation for cross-modality feature prediction[44–49]. Third, multimodal observations can be used to infer the position of a tissue entity in the tissue state space and thereby derive a representation of its state, which goes beyond conventional multimodal integration frameworks that primarily align or fuse information from different modalities. Representations inferred from different modalities of the same entity are expected to be consistent. Increasing the number of modalities, as well as the resolution and depth of measurement, can reduce uncertainty in tissue-state estimation and yield more robust, accurate, and fine-grained tissue representations. Fourth, tissue state representations preserve spatial information and can be decoded into modality-specific spatial features. The decoded features should satisfy cycle-consistency[50], whereby predicted features can be mapped back to the same tissue state representations, providing a way to assess the fidelity of the generated features.

On the basis of these assumptions, AIVT is envisioned to possess three core capabilities (Fig. 1), including tissue state representation, spatial molecular and morphological feature prediction, and simulation of spatiotemporal tissue dynamics.

## Tissue state representation

AIVT aims to provide unified and generalizable representations of tissue states that are robust to the type and number of modalities, measurement platforms, and technical noise. Rather than serving only as modality-aligned embeddings, these representations are intended to capture the underlying biological states of tissue entities while preserving spatial information. They can support a wide range of downstream analysis tasks across scales[51,52], from cellular segmentation, annotation, and lineage assignment, to spatial clustering, tissue segmentation and registration, and tissue-level disease classification, diagnosis, prognosis assessment, and treatment response prediction[53]. In addition, spatial tissue state representations can serve as a decision-making layer for experimental design in spatial omics, such as optimizing tissue sectioning strategies[54] and field-of-view selection[55].

## Molecular and morphological feature prediction

AIVT enables prediction of unmeasured features, including missing molecular layers (e.g., genes and protein biomarkers), spatial features outside captured regions of tissue sections or samples, and unavailable modalities, thereby extending the granularity, scope, and depth of analysis. This capability is particularly valuable when sample availability is limited, experimental costs are high[56], or techniques are destructive, making simultaneous acquisition of multimodal data from the same sample difficult. For example, AIVT could be constructed from morphology data (e.g., hematoxylin and eosin (H&E) images) alone and then used to predict spatial transcriptomic or proteomic features. These predictions can support rapid in silico screening of biomarkers and guide the selection of targets and measurement depth in subsequent wet-lab experiments. Moreover, the cross-modality prediction ability of AIVT implies cycle consistency[50], where predicted features can be used to deduce the input features in a closed loop. This property can help validate the fidelity of results and enhance the interpretability of AIVT.

## Spatiotemporal dynamic simulation

AIVT is designed to support data-driven simulation of spatiotemporal tissue dynamics during development, disease progression[57,4] and responses to perturbations, and enable lineage inference and identification of spatiotemporal feature patterns. By modeling state transition processes and modulating latent tissue state representations, AIVT could estimate tissue responses to genetic[58], pharmacological, or environmental perturbations[59]. It therefore has the potential to serve as an in silico experimental platform for hypothesis generation, mechanism exploration, and experiment optimization, helping researchers identify biologically informative questions and optimize experimental conditions at lower cost.

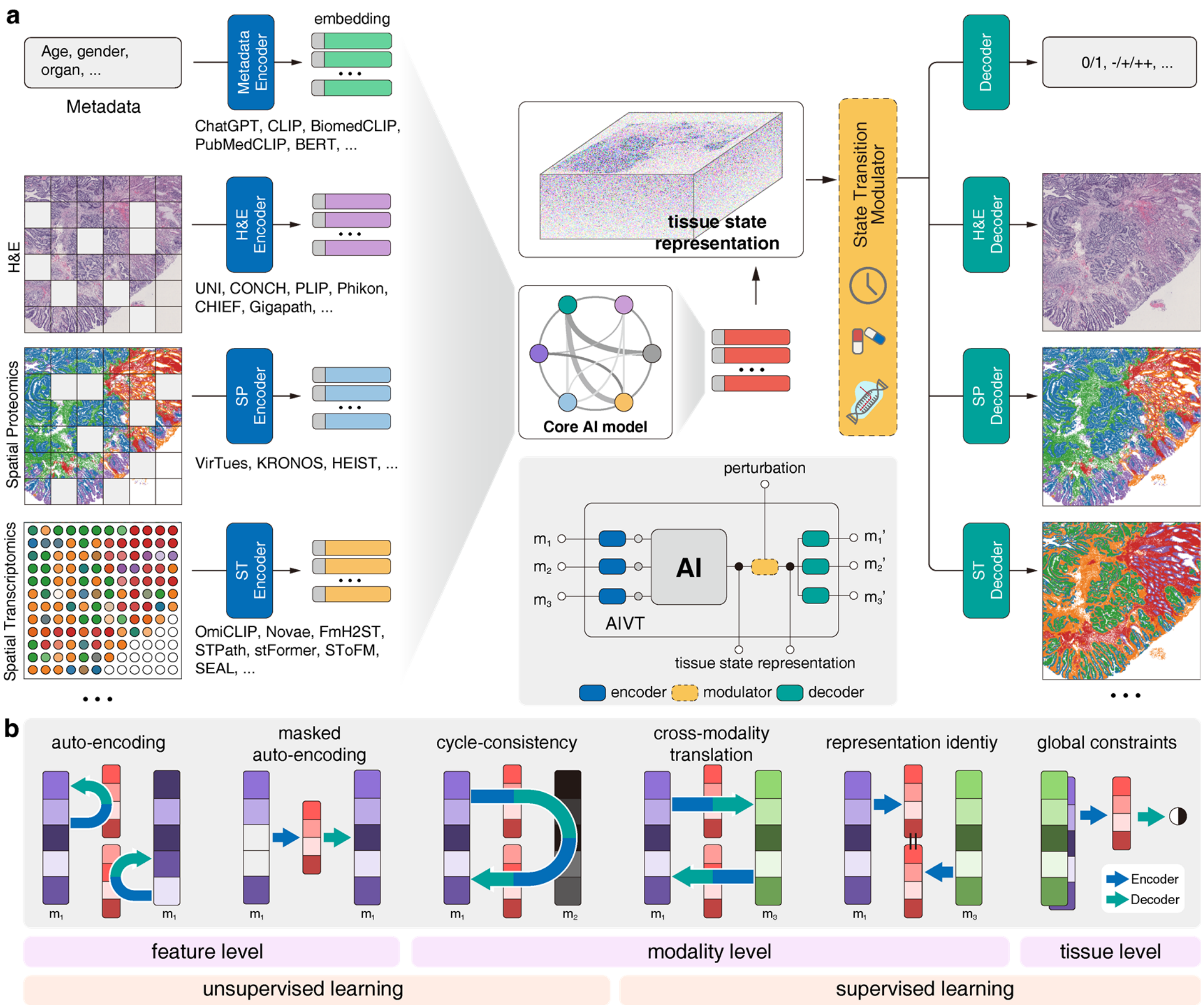

**Fig. 2 | Overview of AIVT. a**, AIVT consists of modality-specific encoders, a core AI model, state transition modulators, and modality-specific decoders. Measurement data from different modalities are encoded into unified numerical embeddings through corresponding encoders. The core AI model utilizes this information to localize tissue states within the latent tissue state space and generates a virtual tissue state representation. State transition modulators model the transition of tissue states under various conditions and modulate the latent representation accordingly. The decoders then map tissue state representations into interpretable outputs. **b**, To realize the proposed assumptions and expected capabilities, AIVT is constrained at different levels, from feature level to modality and tissue levels, through both unsupervised and supervised learning.

## Building the AIVT

Based on the proposed assumptions and capabilities, AIVT is envisioned as a multimodal AI framework composed of four main components: modality-specific encoders, a core AI model, state transition modulators, and modality-specific decoders (Fig. 2a). The modality-specific encoders transform heterogeneous measurement data from different modalities, such as morphological images, molecular profiles, and clinical metadata, into numerical embeddings in a unified format. Recent advances in foundation models provide standardized input interfaces for AIVT across diverse data types. For example, large language models can be used to process textual information related to tissues[60–64], and a wide range of vision and multimodal foundation models can encode morphological images[65–69], spatial transcriptomics[70–76], spatial proteomics[77–79], and other types of data. The core AI model is a multimodal foundation model, which integrates embeddings from different modalities, infers the position of tissue state in the latent tissue state space, and outputs a virtual tissue state representation. The state transition modulators model the state transitions under different conditions, such as treatment, genetic perturbation, environmental change, development, and disease progression, and modulate the tissue state representations accordingly. In this way, AIVT can describe how tissue states shift over time and in response to perturbations, thereby supporting the simulation of spatiotemporal tissue dynamics. The decoders map latent tissue state representations into interpretable outputs, such as reconstructed features in specific modalities and clinically relevant phenotypes. Through these decoders, latent tissue state representations are projected back into explicit and concrete observation domains, making AIVT more interpretable and amenable to validation.

To satisfy the proposed assumptions and realize the intended capabilities, AIVT should be constrained at multiple levels through both unsupervised and supervised learning (Fig. 2b). At the feature level, correlations among features within each modality permit reconstruction-based learning, including auto-encoding[80] and masked auto-encoding[81], through AIVT. At the modality level, AIVT should exhibit several key properties: (1) cross-modality translation, allowing one modality to be predicted from another; (2) cycle consistency[50], where translated outputs can be mapped back to the source modality; (3) representation identity, such that, for the same tissue entity, the tissue state representations inferred from different modalities are consistent. At the tissue level, AIVT should capture global tissue attributes that support tissue-level analysis. Moreover, the combination of unsupervised and supervised constraints broadens the range of usable data, from single-modality observations to paired multimodal measurements.

## Foundations for building AIVT

The rapidly growing volume of spatial multimodal data, coupled with advances in AI algorithms, provides the foundation for building AIVT. Histological images, such as hematoxylin and eosin (H&E) images, have been widely used in routine diagnosis and represent the most abundant resource for characterizing tissue morphology across various species, populations, organs, and diseases. They capture a broad range of nuclear, cytoplasmic, and extracellular matrix features, thereby providing fertile ground for the development of histopathological foundation models[65,67–69,82–84] and serving as a central modality in spatial omics analysis[85]. Advances in spatial omics technologies have further enabled the interrogation of diverse molecular modalities alongside morphological information within shared spatial coordinates. The number of datasets simultaneously profiling two modalities (e.g., HEST-1k[86], STimage-1K4M[87], SpatialDB[88], and STOmicsDB[89] for histology and spatial transcriptomic, and datasets from GigaTIME[90], HEX[47], and VirTues[77] for histology and spatial proteomic) or even more modalities[91] is rapidly increasing. As multimodal datasets continue to expand, AIVT will be able to learn relationships across modalities and represent tissue states with increasing robustness and accuracy. In addition, emerging three-dimensional measurement techniques, such as OTLS microscopy[12,13], microCT[14], DISCO-MS[92], and Open-ST[93], enable a more comprehensive understanding of tissue physiology and pathophysiology, while providing broader spatial context for tissue modeling[94].

The heterogeneity, complexity, and large volume of spatial multimodal data pose substantial challenges for computational modeling. Deep learning has become an invaluable tool for processing spatial data, extracting biologically relevant features, and performing various downstream tasks. Convolutional neural networks and vision transformers[95,96] are the most commonly used architectures in histology to learn spatial feature representations, while graph neural networks are more widely used in spatial omics to model relationships between cells and across modalities[42,97–99]. The recently introduced 'Super Transformer'[100] concept aims to create a unified architecture capable of processing data from different modalities within a single model, thereby supporting more comprehensive tissue modeling. In addition, advances in training strategies can further enhance the generalization and efficiency of the model. For example, self-supervised learning techniques[101] (e.g., contrastive learning, masked modeling, and generative learning) enable more effective use of data from different sources without requiring extensive annotations. Multitask learning[102–105] can improve the generalization and versatility of model, and may facilitate the learning of shared tissue state representations.

## Challenges and outlook

Although existing spatial multimodal data and AI techniques provide a foundation for the development of AIVT, several important challenges remain to be addressed. Current techniques for co-mapping multiple modalities within the same tissue section are typically limited to two or three modalities[26,106,107] and constrained by low spatial resolution or a restricted number of detectable targets[108]. Simultaneous high-resolution measurement of more modalities is necessary for AIVT to accurately model the relationships between molecular features and tissue structure, and to rigorously validate its capabilities. Most currently available datasets are two-dimensional and provide static snapshots of tissues. Three-dimensional measurements would offer a more comprehensive view of tissue state and richer neighborhood context for building AIVT, while longitudinal measurements of morphological and molecular changes within the same tissue are essential for simulating spatiotemporal tissue dynamics. The scale and high dimensionality of the data challenge the scalability and computational efficiency of AIVT, calling for more efficient model architectures and further advances in computational infrastructure. Moreover, model interpretability[109] and the risk of inaccurate predictions remain challenges in the development of AIVT, highlighting the need for more explainable and trustworthy modeling approaches[110].

## Acknowledgements

This work is supported by grants from the National Natural Science Foundation of China (grant 62471212 and 62201221 to S.C.), Guangdong Basic and Applied Basic Research Foundation (2025A1515011333 to S.C.), Science and Technology Projects in Guangzhou (grant 2024A04J4960 to S.C.).

## Author contributions

S.C. and Q.L. conceptualized the project. S.C. supervised the overall research. Q.L. and S.C. drafted the initial manuscript and prepared the figures. S.C., S.Z., and Q.F. refined the concepts and methodology. All authors reviewed, edited and approved the final manuscript.

## Competing interests

The authors declare no competing interests.